\documentclass[11 pt]{article}
\usepackage[utf8]{inputenc}
\usepackage{url}
\usepackage{authblk}
\usepackage{amsmath}
\usepackage{amssymb}
\usepackage{hyperref}
\usepackage{graphics}
\usepackage{mathtools}
\usepackage{caption}
\usepackage{subcaption}

\usepackage{graphicx}
\usepackage{algpseudocode}
\usepackage{algorithm}
\usepackage{afterpage}
\usepackage{array}
\usepackage{collcell}
\usepackage{multirow}
\usepackage{amsthm}

\usepackage{algpseudocode}
\usepackage{algorithm}

\def\w{{\boldsymbol{\omega}}}

\def\A{{\mathbf A}}
\def\c{{\mathbf c}}

\newcommand{\LinesNumbered}{
  \setboolean{algocf@linesnumbered}{true}
  \renewcommand{\algocf@linesnumbered}{\everypar={\nl}}}
\let\oldnl\nl
\newcommand{\nonl}{\renewcommand{\nl}{\let\nl\oldnl}}

\usepackage{algorithm,algpseudocode}

\newcommand{\argmin}[1]{\underset{#1}{\mathrm{argmin}}}

\newtheorem{theorem}{Theorem}[section]

\def\c{{\mathbf c}}

   \def\A{{\mathbf{A}}}

\usepackage{caption} 
\usepackage{multirow}

\newcommand{\RR}{\mathbb{R}}

\usepackage[left=1in,right=1in,top=1.2in,bottom=1.2in,
            footskip=.25in]{geometry}

\begin{document}
\title{SPADE4: Sparsity and Delay Embedding based Forecasting of Epidemics}

\author[1]{Esha Saha\footnote{The code is available at \url{https://github.com/esha-saha/spade4}.}}
\author[2]{Lam Si Tung Ho} 
\author[1]{Giang Tran}

\affil[1]{Department of Applied Mathematics, University of Waterloo}
\affil[2]{Department of Mathematics and Statistics, Dalhousie University}

\date{ }

\maketitle
\begin{abstract}
    Predicting the evolution of diseases is challenging, especially when the data availability is scarce and incomplete. The most popular tools for modelling and predicting infectious disease epidemics are compartmental models. They stratify the population into compartments according to health status and model the dynamics of these compartments using dynamical systems. However, these predefined systems may not capture the true dynamics of the epidemic due to the complexity of the disease transmission and human interactions. In order to overcome this drawback, we propose \textbf{Spa}rsity and \textbf{D}elay \textbf{E}mbedding based \textbf{Fore}casting (SPADE4) for predicting epidemics. SPADE4 predicts the future trajectory of an observable variable without the knowledge of the other variables or the underlying system. We use random features model with sparse regression to handle the data scarcity issue and employ Takens' delay embedding theorem to capture the nature of the underlying system from the observed variable. We show that our approach outperforms compartmental models when applied to both simulated and real data. 
\end{abstract}


\section{Introduction}
Humans and pathogens have always been in a race to beat each other. Whenever the human community is introduced to new diseases, people either control the spread through various measures or gain immunity through infection, medications and/or vaccinations. However, when a disease is new to the community, there is a need to understand, analyze and predict its evolution so that at a social level, spread of the disease can be managed and enough information regarding the disease causing pathogen can be made available for scientists to come up with medications and/or vaccinations. Numerous works have been trying to study the nature of epidemics using mathematical techniques in epidemiology and/or machine learning.

Surveillance data, such as daily cases, are the prime source of information about emerging infectious disease epidemics.
Predicting the trajectories of epidemics from surveillance data is currently one of the most active research areas, partially due to the COVID-19 pandemic.
Compartmental models, which stratify the population into compartments according to health status, are the most popular tools for this task.
They have been used to study many infectious disease epidemics including plague \cite{ho2018birth}, Ebola \cite{althaus2014estimating, ho2018direct}, measles \cite{cauchemez2008likelihood}, HIV \cite{blum2010hiv}, influenza \cite{dukic2012tracking}, and COVID-19 \cite{sueir, cramer2022evaluation}.
These models utilize dynamical systems to describe the dynamics of their compartments.

For example, the classic Susceptible-Exposed-Infectious-Recovered (SEIR) divides the population into four groups: Susceptible (healthy), Exposed (infected but not yet infectious), Infectious (infected and infectious), and Recovered (recovered or died).
Let $S, E, I, R$ be the population of Susceptible, Exposed, Infected, Recovered groups, respectively.
Under the SEIR model, we have
\begin{equation}\label{eqn:SEIR}
    \begin{aligned}
    \dfrac{dS}{dt} & = -\dfrac{\beta S I}{P}\\
    \dfrac{dE}{dt} &= \dfrac{\beta S I}{P} - \sigma E\\
    \dfrac{dI}{dt} & = \sigma E - \gamma I\\
    \dfrac{dR}{dt} &= \gamma I,
    \end{aligned}
\end{equation}
where $\beta$ is infection rate, $\sigma$ is the latency rate, $\gamma$ is the removal rate, and $P$ denotes the population. Note that the parameter $\beta$ may be a function of time based on the assumption that infection rate varies over the epidemic cycle \cite{smirnova2019forecasting,girardi2023seir}. By learning these parameters from surveillance data, one can predict the future trajectories of the epidemic.
However, the SEIR model is often too simple to fully capture the dynamics of an epidemic.
Therefore, much effort has been made to modify the SEIR model according to the nature of the infectious disease.
One such model is the S$\mu$EIR model described in \cite{sueir} which takes into account unreported cases. The dynamics of this model is given by
\begin{equation}\label{eqn:SuEIR}
    \begin{aligned}
    \dfrac{dS}{dt} & = -\dfrac{\beta (I+E)S}{P}\\
    \dfrac{dE}{dt} &= \dfrac{\beta(I+E)S}{P} -\sigma E\\
    \dfrac{dI}{dt} & = \mu\sigma E - \gamma I\\
    \dfrac{dR}{dt} &= \gamma I,
    \end{aligned}
\end{equation}
where the parameter $\mu$ represents the discovery rate which characterizes the ratio of the exposed cases that are confirmed and reported to the public. It reflects the unreported/undiscovered cases. Nevertheless, the performance of compartmental models depends heavily on how well the predefined dynamical system can capture the dynamic of an epidemic.
In this paper, we propose a machine learning method for predicting infectious disease epidemics.
Unlike compartmental models, our approach does not make any assumption on the underlying dynamical system.

Typical machine learning or artificial intelligence based methods aim to learn the trajectory of a dynamical system from a $d-$dimensional time dependent data $\mathbf{x}(t)\in\mathbb{R}^d$ by finding a function $f$ such that $\dfrac{d\mathbf{x}(t)}{dt} = f(\mathbf{x}(t))$. 
This comes under an active area of research involving learning of dynamical systems from data. One class of methods assumes the underlying dynamical system to have sparse representations with respect to a prescribed dictionary consisting of polynomials and/or trigonometric functions. Sparse regression is used to identify these active terms \cite{kaiser2018sparse, mangan2017model,sindy,rudy2017data, tran2017exact, schaeffer2018extracting,schaeffer2017learning,goyal2022discovery}. Another approach based on neural networks is modeling partial or ordinary differential equations using either full data or partial observations with some side information to learn the system \cite{weinan2017proposal,narendra1992neural,partialobs,pinn, gonzalez1998identification,lagaris1998artificial,sitzmann2020implicit,lu2021deepxde,lusch2018deep,qin2019data, su2021deep,vlachas2018data}. Other works in \cite{lu2019deeponet,li2020fourier,bhattacharya2020model,dongbin} demonstrate the power of using deep learning for approximating operators between functional spaces. However, neural networks perform well only when they have enough training data. This makes their applications limited in the data scarce regime. Another issue with neural networks is the lack of theoretical understanding owing to their complexity. Random feature models proposed in \cite{rahimi2007random,rahimi2008uniform,rahimi2008weighted} provide an alternative to using neural networks without compromising on the accuracy. Results in \cite{jacot2020implicit,yang2021exact,weinan2019comparative} demonstrate that it is possible to gain theoretical understanding of random feature models. These classes of methods have also been successfully used in \cite{nelsen2021random} for learning operators between Banach spaces. Sparse Random Feature Extraction proposed in \cite{hashemi2021generalization} modifies random feature models to make it more suitable for data scarce problems by incorporating sparsity in building of the feature space. Similar works on sparse random feature models and feature selection using different methods have been widely explored in \cite{saha2022harfe, srmd, shrimp}. Most of those methods rely on the assumption that full state information is available. However, epidemiological data are often only partially observed where available observations are either daily active cases or cumulative cases. 
Therefore, the machine learning methods mentioned above are not applicable for learning the dynamics of epidemics. 
Motivated by various delay embedding theorems such as the Whitney embedding theorem and Takens' theorem \cite{takens1981detecting}, we employ the delayed embedding method in our epidemiological forecast model. Delay embedding has been used to develop eigensystem realization algorithm in system identification to learn the eigensystem \cite{juang1985eigensystem}, to extract qualitative dynamics from data \cite{broomhead1986extracting}, and to study nonlinear dynamics in the Koopman operator framework \cite{le2017higher, brunton2017chaos, champion2019discovery, kamb2020time}. It can also be combined with neural networks to identify parameters and make prediction \cite{lin2022parameter, uribarri2022dynamical}. It is worth noting that Takens' embedding theorem identifies the model up to a diffeomorphism. 
Therefore, learning the original dynamical system from partially observed using delay embedding remains open.

In our work, we illustrate that it is feasible to forecast future dynamics of infectious disease epidemics using delay embedding. Specifically, we approximate the rate of change in the observed variable (daily active or cumulative cases) as a sparse random feature expansion of its time delayed mapping. Here, a sparsity constraint in the random feature model helps balance between the richness representation of the function space and the limited amount of available data. 
Then, we use the learned function to make a prediction over a future time window. 
The proposed procedure is called Sparsity and Delay Embedding based Forecasting (SPADE4).
We test SPADE4 on simulated data and consider cases where the input is without or with noise, and also test with different levels of noise. We show that SPADE4 performs consistently well over the SEIR and S$\mu$EIR models on simulated data.
Moreover, SPADE4 outperforms these models in predicting future trajectories of various epidemics including COVID-19 in Canada, Ebola in Guinea, Zika in Giradot, and Flu in China.

\section{Motivation and Problem Setting}
Throughout the paper, we use bold letters for column vectors (e.g. $\mathbf{x}$) and bold capital letters for matrices (e.g. $\mathbf{A}$). A dot above a variable denotes time differentiation (e.g. $\dot{y}(t)$). We say that a vector $\mathbf{z}$ is $s$-sparse if it has at most $s$ nonzero entries.  Let $[N]$ denote the set of all positive integers less than or equal to $N$. 

Given time-dependent observations $\{y(t_k)\}_{k=1}^m$ from an unknown multidimensional dynamical system, we propose a new inference method motivated by Takens' Theorem where we aim to forecast the values of $y(t)$ over a given forecast horizon (generally one-week-ahead forecast). First proposed in 1981 in \cite{takens1981detecting}, given a dynamical system $\varphi:M\rightarrow M$, ($M$ is a compact manifold of dimension $d$) and an observable $y:M\rightarrow\mathbb{R}$, Takens' theorem aims to obtain information about the original dynamical system. We recall the formal statement of Takens' Theorem as stated in \cite{takens1981detecting, huke2006embedding}. 

\begin{theorem}[Takens' Theorem]\label{thm:takens}. Let $M$ be a compact manifold of dimension $d$. For pairs $(\varphi,y)$, with $\varphi:M\rightarrow M$ a smooth diffeomorphism and $y:M\rightarrow\mathbb{R}$ a smooth function, it is generic property that the map $\Phi_{(\varphi,y)}:M\rightarrow \mathbb{R}^{2d+1}$, defined by
\begin{equation}\label{eq: takens}
    \Phi_{(\varphi,y)}(\mathbf{x}) = (y(\mathbf{x}),y(\varphi(\mathbf{x})),...,y(\varphi^{2d})(\mathbf{x})))
\end{equation}
is an embedding.
\end{theorem}
The term `generic' here means open and dense and by ``smooth" we mean at least twice continuously differentiable and the function $\Phi$ is called the time delay mapping. 

Motivated by Takens' theorem, we assume that the rate of change in the observable $y(t)$ is a function of its time delayed mapping $\Phi$, i.e.,
\begin{equation}\label{eq:dxdt=f}
    \dot{y}(t_k) = f(y(t_k), y(t_{k-1}),...,y(t_{k-(p-1)})),
\end{equation}
where $p$ is the embedding dimension. Note that the variable $y$ is part of an unknown multidimensional dynamical system. We would like to forecast the value of $y(t)$ while preserving the properties of the corresponding multidimensional dynamical system. Takens' theorem provides a theoretical guarantee to preserve the topological structure of the multidimensional dynamical system while solving the delayed equation, under certain conditions of the dynamics. Specifically, we wish to learn the function $f:\mathbb{R}^p\rightarrow\mathbb{R}$ in Equation \eqref{eq:dxdt=f} of the form
\begin{equation}\label{f.approx}
f(\mathbf{h}) \approx \sum\limits_{j=1}^N c_j \phi(\langle \mathbf{h},\boldsymbol{\omega}_j\rangle),
\end{equation} 
where $\boldsymbol{\omega}_j\in\mathbb{R}^{p}$ are the random weights, $\phi$ is a nonlinear activation function, and $\c =\begin{bmatrix}c_1\ldots c_N\end{bmatrix}^T\in\RR^N$ is the trainable coefficient vector. The nonlinear activation function $\phi: \RR \rightarrow \RR$ can be chosen to be a trigonometric function, the sigmoid function, or the ReLU function. Here, we use the ReLU activation function, i.e. $\phi(\cdot)=\max\{0,\cdot\}$. Entries of the random weight vector are i.i.d. random variables generated by a probability function $\rho(\omega)$, while the coefficients $\c\in\RR^N$ are trainable. Also known as random features model proposed in \cite{rahimi2007random,rahimi2008uniform,rahimi2008weighted}, this model can be considered as a wide two-layer neural network where the weights in the first hidden layer is generated (following a distribution) and frozen while training problem relies on learning the coefficient vector $\c$. Since we assume limited data availability, we wish to have a sparse representation of the function $f$ by learning the coefficient vector $\mathbf{c}$ with a sparsity constraint.  Theoretically, when $N$ is very large, the random feature methods has been shown to be comparable with shallow networks in terms of theoretical risk bounds \cite{rahimi2007random, rahimi2008weighted, rahimi2008uniform, rudi2017generalization}. 

Given $m$ measurements of the obervable $y$, we first build the input-output pairs $\{(\mathbf{h}_k,\dot{y}(t_k))\}_{k=p}^m$, where $\mathbf{h}_k= [y(t_k),y(t_{k-1}),...,y(t_{k-p+1)})]^T$ for $k=p,p+1,..,m$. We approximate the output data $\{\dot{y}(t_k)\}_{k=1}^m$ from given input data $\{y(t_k)\}_{k=1}^m$ using finite difference methods. In all our numerical simulations, we build the output data as follows:
\begin{equation}
\begin{aligned}
    \dot{y}(t_1) &= \dfrac{y(t_2)-y(t_1)}{(t_2-t_1)}; \\ 
    \dot{y}(t_k) &= \dfrac{y(t_{k+1})-y(t_{k-1})}{(t_{k+1}-t_{k-1})}, \quad \text{ for } k=2,..,m-1; \\
    \dot{y}(t_m) &= \dfrac{y(t_m)-y(t_{m-1})}{(t_m-t_{m-1})}.
    \end{aligned}
\end{equation}

\begin{figure}
    \centering
    \includegraphics[width = 1\textwidth]{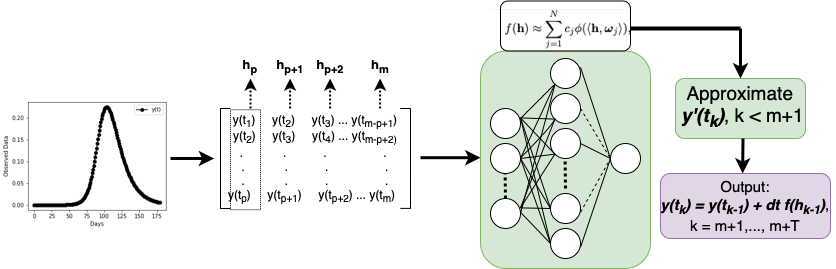}\\
    \caption{Schematic representation of SPADE4 algorithm.}
    \label{fig:flowchart}
\end{figure}

We used central Euler discretizations for the intermediate points to have second order accuracy and minimize the errors arising from using discrete methods. Other derivative approximation techniques could also be used, such as forward or backward discretizations. If the data is noisy, then an additional step would be required after getting the output data since finite difference methods may amplify the noise in the input dataset making recovery difficult. We apply a convolution based averaging filter on $\{\dot{y}(t_k)\}_{k=1}^m$ with a smoothing parameter $s$ as given below,
\begin{equation}\label{eq:smooth}
    \dot{y}(t_k) = \dfrac{1}{s}\sum\limits_{n = (k+1)-(s-1)}^{k+1}\dot{y}(t_n),
\end{equation}
where $\dot{y}(t_k) = 0\,\,\forall\,\, k\leq 0 \text{ and } k\geq m+1$ and $s$ denotes the strength of the smoothing filter. The value of $s$ chosen is dependent upon the noise level present in the dataset. The value of $s$ used for each of the experiments have been specified in their respective sections. Let $\mathbf{z} = [\dot{y}(t_{p}),\dot{y}(t_{p+1}),...,\dot{y}(t_m)]^T$ and $\mathbf{A} = (\phi(\langle \mathbf{h}_k,\boldsymbol{\omega}_j\rangle))\in\mathbb{R}^{ (m-p+1)\times N}$. The matrix $\mathbf{A}$ is given by:
\begin{equation}\label{eq:matA}
    \mathbf{A} = \begin{bmatrix}
    \phi(\langle \mathbf{h}_p,\boldsymbol{\omega}_1\rangle) & \phi(\langle \mathbf{h}_{p},\boldsymbol{\omega}_2\rangle)& \phi(\langle \mathbf{h}_{p},\boldsymbol{\omega}_3\rangle) & \hdots & \phi(\langle \mathbf{h}_p,\boldsymbol{\omega}_N\rangle)\\
   \phi(\langle \mathbf{h}_{p+1},\boldsymbol{\omega}_1\rangle) & \phi(\langle \mathbf{h}_{p+1},\boldsymbol{\omega}_2\rangle) &\phi(\langle \mathbf{h}_{p+1},\boldsymbol{\omega}_3\rangle)&\hdots &\phi(\langle \mathbf{h}_{p+1},\boldsymbol{\omega}_N\rangle) \\
    \vdots & \vdots & \vdots & \vdots & \vdots \\ 
    \phi(\langle \mathbf{h}_m,\boldsymbol{\omega}_1\rangle) & \phi(\langle \mathbf{h}_{m},\boldsymbol{\omega}_2\rangle) & \phi(\langle \mathbf{h}_{m},\boldsymbol{\omega}_3\rangle) & \hdots & \phi(\langle \mathbf{h}_{m},\boldsymbol{\omega}_N\rangle)
    \end{bmatrix}\in\mathbb{R}^{ (m-p+1)\times N}
\end{equation}
The problem \eqref{f.approx} becomes,
\begin{equation}\label{eq:opt}
    \text{find } \mathbf{c}\in \RR^N\  \text{such that } \mathbf{z}\approx A\mathbf{c} \quad \text{and}\quad \c \quad \text{is sparse},
\end{equation}
which can be solved by the following minimization problem:
 \begin{equation}\label{eqn:lasso}
   \mathbf{c}^\# = \argmin{\mathbf{c}\in\RR^N} \|\A\mathbf{c} - \mathbf{z}\|_2^2 + \lambda \|\mathbf{c}\|_1\quad 
\end{equation}
Here, $\lambda>0$ is the regularization parameter. To forecast $T$ future values of the given trajectory i.e., $\{y(t_{m+i})\}_{i=1}^T$, we use the Euler method: 
\begin{equation}\label{eqn:forecastI}
    y(t_{m+i}) = y(t_{m+i-1}) +  (t_{m+i} - t_{m+i-1})f(\mathbf{h}_{m+i-1}), 
\end{equation}
for all $i=1,...,T$ and $\mathbf{h}_k = [y(t_k),y(t_{k-1}),...,y(t_{k-p+1)})]^T$ for any integer $k = m,m+1,...m+T$. The summary of our SPADE4 algorithm is given in Algorithm \ref{alg:main} and schematically represented in Figure \ref{fig:flowchart}.

\begin{algorithm}[ht!]
\caption{Algorithm for SPADE4}\label{alg:main}
\begin{algorithmic}[1]

\Require Input data $\{y(t_k)\}_{k=1}^m$; smoothing parameter $s$; function $\phi(\cdot,\boldsymbol{\omega}) =\phi(\langle \cdot,\boldsymbol{\omega}\rangle)$, distribution $\rho$ to sample $\boldsymbol{\omega}$, number of features $N$, regularization parameter $\lambda$. 
\Ensure
\State{Approximate $\{\dot{y}(t_k)\}_{k=1}^m$: } 
\[
\begin{aligned}
    \dot{y}(t_1)&\gets \dfrac{y(t_2)-y(t_1)}{(t_2-t_1)},\quad \dot{y}(t_m)\gets \dfrac{y(t_m)-y(t_{m-1})}{(t_m-t_{m-1})},\\
\dot{y}(t_k)&\gets \dfrac{y(t_{k+1})-y(t_{k-1})}{(t_{k+1}-t_{k-1})},\quad \text{for}\quad k=2,3,\ldots,m-1.
\end{aligned}
\]

\If{$\{y(t_k)\}_{k=1}^m$ is noisy}
 \[\dot{y}(t_k) \gets \dfrac{1}{s}\sum\limits_{n = k+2-s}^{k+1}\dot{y}(t_n)\]
\Statex{\text{ where }} $\dot{y}(t_k) = 0\,\,\forall\,\, k\leq 0 \text{ and } k\geq m+1$.
\EndIf

\medskip

\State{Define $\mathbf{z} = [\dot{y}(t_p),\dot{y}(t_{p+1}),...,\dot{y}(t_m)]^T$.}

\medskip

\State{Draw $N$ random weights $\w_j\sim \rho(\w)$.}

\medskip

\State{Define the time-delay data \[\mathbf{h}_k = [y(t_k),y(t_{k-1}),...,y(t_{k-p+1)})]^T,\quad\text{for}\quad k=p,\ldots,m.\] }

\State{Construct random feature matrix $\mathbf{A}= (\phi(\langle\mathbf{h}_{k};\w_j\rangle)) \in \RR^{(m-p+1) \times N}$.}

\medskip

\State{Solve
\[ \mathbf{c}^\# = \argmin{\mathbf{c}\in\RR^N} \|\A\mathbf{c} - \mathbf{z}\|_2^2 + \lambda \|\mathbf{c}\|_1. \]
 }
 \For {$i=1,\ldots,T$}
 \State{
$\hat{f}(\mathbf{h}_{m+i-1}) = \sum\limits_{j=1}^N c_j^\#\phi(\langle\mathbf{h}_{m+i-1};\boldsymbol{\omega}_j\rangle)$}
\State{$y(t_{m+i}) = y(t_{m+i-1}) + (t_{m+i} - t_{m+i-1})\hat{f}(\mathbf{h}_{m+i-1}).$}
\EndFor

\medskip

 \State {\bf Output:} $\{y(t_k)\}_{k=m+1}^{m+T}$.
\end{algorithmic}
\end{algorithm}

\section{Numerical Experiments on Synthetic Data}\label{sec:synthetic-data}
In this section, we compare our proposed method and relevant benchmark methods, including the popular SEIR model and the most state-of-the art S$\mu$EIR model \cite{sueir} on synthetic data simulated from the S$\mu$EIR model. In this simulation study, the input data consists of time series $\{I(t_k)\}_{k=1}^m$ corresponding to the infectious variable $I(t)$ in the S$\mu$EIR model. Our goal is to obtain a one-week-ahead forecast horizon of this infectious variable. In all experiments, we display the predicted values of the mentioned methods versus the ground truth when varying the size of the training data as the observed trends change (before the peak, after the peak, and around the peak of the epidemic). We also study the robustness of SPADE4 with respect to noise and compare with the benchmark models.

\medskip

\noindent{\bf Data Generation.} We first solve the S$\mu$EIR model given in Eq. \eqref{eqn:SuEIR} numerically on the time interval $[0,180]$ (days) with the timestep $\Delta t=0.01$ and the parameters of the S$\mu$EIR model:  
\[\beta = 3/14,\quad  \sigma =0.25 ,\quad \mu=0.75,\quad \gamma =1/14,\quad P=S(0)+E(0)+I(0)+R(0),\]
and initial conditions are 
\[S(0)= 10^6, \quad E(0) =0,\quad I(0) =1,\quad R(0) =0.\] 
We normalize the solution by the population $P$ before building the training dataset.
The training dataset is either noiseless $\{I(t_k)\}_{k=1}^m$, or noised $\{I(t_k)+ \varepsilon_k\}_{k=1}^m$, which resembles the reported daily counts of active cases in an outbreak. Note that the inputs are proportions instead of counts in this study. The size of the training data $m$ will be specified in each experiment.

\medskip

\noindent{\bf Prediction using SPADE4.}
For our proposed method, the random weights $\boldsymbol{\omega}_j\in \RR^{p}$ are sampled from $\mathcal{N}(0,1)$ and bias terms $b_j (j\in [N])$ are sampled from $\mathcal{U}(0,2\pi)$. The number of random weights is $N=50m$, where $m$ is the number of training points. The activation function is set to $\phi(\cdot)=\text{ReLU}(\cdot)$ and the time delay $\tau = 1$. We choose the embedding dimension $p = 2d +1$, where $d=4$ is the dimension of the multi-dimensional dynamical system. The $\ell_1$-regularization parameter $\lambda$ is selected by the model from a range of possible values, $\lambda\in [10^{-6},5\times 10^{-6},10^{-7},5\times 10^{-7},10^{-8},5\times 10^{-8},10^{-9},5\times 10^{-9}]$ using the Bayesian Information Criterion (BIC). Using the training data, we obtain the learned coefficients $\c$ by solving the optimization problem \eqref{eqn:lasso}. In our paper, we choose the LASSO package in Python for obtaining $\mathbf{c}$. We forecast the proportion of daily active cases denoted by $\{\hat{I}_{ours}(t_k)\}_{k=m+1}^{m+T}$ ($T$ denotes the prediction window) using Equation \eqref{eqn:forecastI}.

\medskip
\noindent{\bf Prediction using benchmark models.} 
For benchmark models, we use least squares method to learn the models' parameters (from 100 possible random initializations) by fitting given input data with the variable $I(t)$ from S$\mu$EIR and SEIR models as the target function. The parameters learnt are $(\beta, \sigma,\mu,\gamma)$ for the S$\mu$EIR model (Equation \eqref{eqn:SuEIR}) and $(\beta,\sigma,\gamma)$ for SEIR model 
(Equation \eqref{eqn:SEIR}).

\medskip

\noindent {\bf Prediction accuracy.} For all the methods, accuracy of predicting the trajectory of a novel pathogen outbreak (such as COVID-19) is measured using relative test error which is calculated as
\begin{equation}
   \text{ Error}(I, \hat{I}_{model}) = \sqrt{\dfrac{\sum\limits_{k=m+1}^{m+7}[ I(t_k) - \hat{I}_{model}(t_k) ]^2}{\sum\limits_{k=m+1}^{m+7} [I(t_k)]^2}},
\end{equation}
where $I$ denotes the true values and $\hat{I}_{model}$ are the predicted values of a model.

\subsection{Results on Noiseless Simulated Data }\label{sec:simul-clean}

\begin{figure}[h!]
    \centering
    
     \begin{subfigure}[b]{1.\textwidth}
     \centering
    \includegraphics[width = 0.3\textwidth]{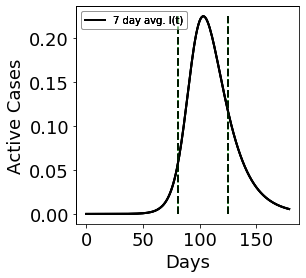}
    \includegraphics[width = 0.3\textwidth]{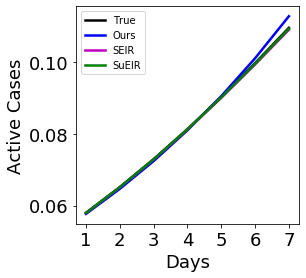}
    \includegraphics[width = 0.3\textwidth]{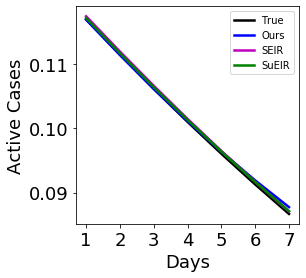}
  
    \end{subfigure}
    \hfill
    \begin{subfigure}[b]{0.9\textwidth}
     \centering
    \includegraphics[width = 0.3\textwidth]{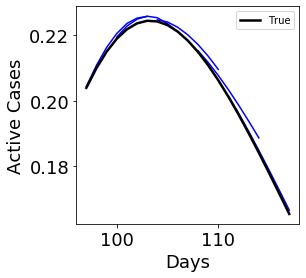}
    \includegraphics[width = 0.3\textwidth]{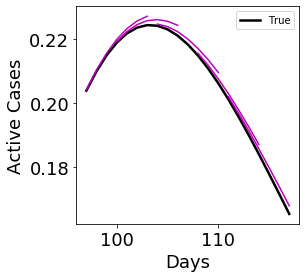}
    \includegraphics[width = 0.3\textwidth]{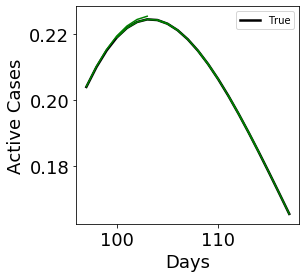}
    \end{subfigure}
     \caption{Results on noiseless simulated data. First row: Noiseless training datasets from 0 upto $81^{st}$ and $125^{th}$ days out of 180 days (top left). The next two figures are the predicted values of the infectious variable $I(t)$ in the next seven days using SPADE4 (blue), SEIR model (magenta), and S$\mu$EIR model (green) correspond to those two training datasets versus ground truth (black). Second row: Prediction of $I(t)$ for next seven days around the peak of the wave using SPADE4 (blue), SEIR model (magenta), and S$\mu$EIR model (green) versus ground truth (black).}
    \label{fig:train-val-SimulNoiselessBef}
\end{figure}

We first present numerical results on simulated data without noise where we evaluate model performance during key moments of the epidemic, i.e., before, around, and after the peak of the curve. For before and after the peak, we consider data points upto 81$^{st}$ and 125$^{th}$ days (given by the dashed vertical lines in the top left plot of Figure \ref{fig:train-val-SimulNoiselessBef}) out of 180 days, respectively, as the training set, and predict for the next seven days. The corresponding predicted curves of the one-week-ahead-forecast horizon are shown in the middle first row and the top right plot of Figure \ref{fig:train-val-SimulNoiselessBef}. For assessing model performance around the changing slope of the trajectory, we consider the number of training points $m\in\{97,100,104,108,111\}$ and plot the predicted curves of our proposed method and the two benchmark models for the one-week-ahead forecast in the second row of Figure \ref{fig:train-val-SimulNoiselessBef}. We observe from Figure \ref{fig:train-val-SimulNoiselessBef} that the best performance is given by the benchmark model fitted with S$\mu$EIR (green curves), closely by our proposed method (blue curves) and the SEIR model (magenta curves). This is an expected outcome considering that we are fitting the benchmark model with the same model from which data was simulated. However, in real world applications, there are irregularities in data which can make the noiseless case too idealistic. So in the following section, we add noise to the input data and then compare the performance of benchmark models to our SPADE4.

\subsection{Results on Noisy Simulated  Data}\label{sec:simul-noise}

\begin{figure}[ht!]
\centering
     \includegraphics[width = 0.3\textwidth]{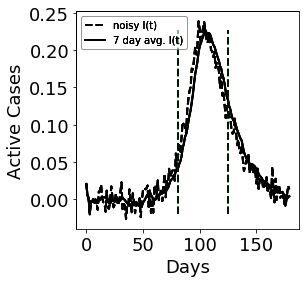}
    \includegraphics[width = 0.3\textwidth]{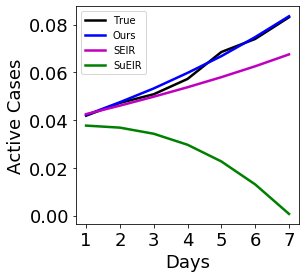}
    \includegraphics[width = 0.3\textwidth]{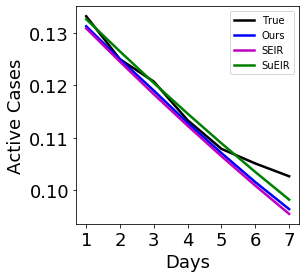}
    \includegraphics[width = 0.3\textwidth]{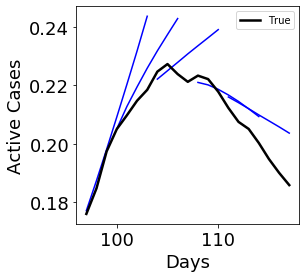}
    \includegraphics[width = 0.3\textwidth]{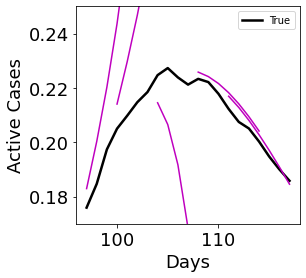}
    \includegraphics[width = 0.3\textwidth]{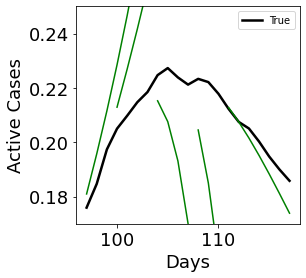}
    \caption{Results on simulated data with $5\%$ noise added to input data. First row: Noisy training datasets from 0 upto $81^{st}$ and $125^{th}$ days out of 180 days (top left). The next two figures are the predicted values of the infectious variable $I(t)$ in the next seven days using SPADE4 (blue), SEIR model (magenta), and S$\mu$EIR model (green) correspond to those two training datasets versus ground truth (black). Second row: Prediction of $I(t)$ for next seven days around the peak of the wave using SPADE4 (blue), SEIR model (magenta), and S$\mu$EIR model (green) versus ground truth (black).  }
    \label{fig:SimulNoisy5}
\end{figure}

In this section we consider noisy input data. As before, the data is simulated from Equation \eqref{eqn:SuEIR} with the parameters described above. The input data with noise is given by:
\begin{equation}
    I_{\text{noisy}}(t) = I(t) + \varepsilon\max\mid{I(t)}\mid,
    \label{eq: noise}
\end{equation}
where $I(t)$ is the clean data, $\varepsilon\sim\mathcal{N}(0,\eta)$, and $\eta$ is the noise level. For all experiments involving noisy input data, we pre-process the input data by considering seven-day average of $I(t)$. This is done to take care of data irregularities and to resemble setup in practice. For our SPADE4, an additional convolution based smoothing filter with parameter $s = 15$ is also used on the output vector $\{\dot{I}(t_k)\}_{k=1}^m$. This is done to minimize noise amplification from the use of finite difference approximations to obtain $\{\dot{I}(t_k)\}_{k=1}^m$. Similar to the setup in Section \ref{sec:simul-clean}, we consider the number of training points $m\in\{81,97,100,104,108,111,125\}$ days and plot the one-week-ahead forecast. We consider $5\%$ noise level, i.e., $\eta = 0.05$ and plot the results in Figure \ref{fig:SimulNoisy5}. Experimental results with $\eta = 0.02$ are given in the appendix.

Before the peak, the one-week-ahead forecast of our SPADE4 outperforms the SEIR and the S$\mu$EIR model in both cases of the noise level. After the peak, the prediction results of SPADE4 and the SEIR model are comparable and are better than those obtained from the S$\mu$EIR (see top row of Figure \ref{fig:SimulNoisy5}). Around the peak, (see bottom row of Figure \ref{fig:SimulNoisy5}), we observe that SPADE4 quickly picks up the changing slope in the wave. On the other hand, the benchmark models give unreliable results at some points in the curve. Both benchmark models overpredict the number of active cases for $m\in\{97,100\}$ and underpredict the number of active cases for $m=104$.

\begin{figure}[ht!]
    \centering
     \includegraphics[width = 0.45\textwidth]{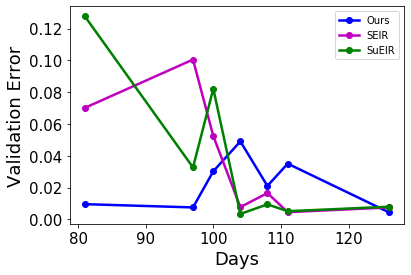}
    \includegraphics[width = 0.45\textwidth]{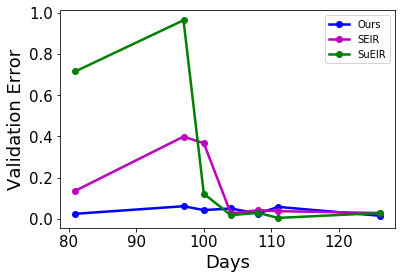}
    \caption{Validation errors based on various size of the training data $m\in\{81,97,101,104,108,112,126\}$ for the noise level $\eta = 0.02$ (left) and $\eta = 0.05$ (right).}
    \label{fig:valerrorsimul}
\end{figure}

The superior performance of SPADE4 can be seen more clearly from Figure \ref{fig:valerrorsimul} where we plot the validation errors for our model and the benchmark models fitted different sizes of training data $m$ and noise levels $\eta = 0.02$ and $\eta = 0.05$. We see that SPADE4 has low relative error consistently while the errors for the other models tend to fluctuate to a higher error at certain points. The performance of the benchmark models is highly influenced by the number of training points, noise level, as well as the nature of the curve while our proposed SPADE4 is robust to all these factors affecting input data.

\section{Numerical Experiments on Real Datasets}\label{sec:real-datasets}
In this section, we demonstrate the performance of SPADE4 on various real datasets, including daily active cases of 2019 Coronavirus (COVID-19) in Canada\footnote{\url{https://health-infobase.canada.ca/covid-19/epidemiological-summary-covid-19-cases.html}} as well as cumulative cases of Ebola in Guinea \footnote{\url{https://www.kaggle.com/datasets/imdevskp/ebola-outbreak-20142016-complete-dataset}}, Zika virus in Giradot \cite{rojas2016epidemiology}, and influenza A/H7N9 in China \footnote{\url{https://datadryad.org/stash/dataset/doi:10.5061/dryad.2g43n}}. The training data for the benchmark models is given by $I(t)$ for the COVID-19 dataset and by the variable $I(t) + R(t)$ for the remaining datasets. For each dataset, we compare the one-week-ahead forecasts of SPADE4 with benchmark methods including SEIR and S$\mu$EIR models across important moments in the epidemic. Predictions using SPADE4 use the same hyperparameters as described in \ref{sec:synthetic-data} unless specified otherwise.

\subsection{Data from Daily Active Cases of COVID-19 in Canada}\label{sec:covid}
\begin{figure}[ht!]
    \centering
    \includegraphics[width = 0.5\textwidth]{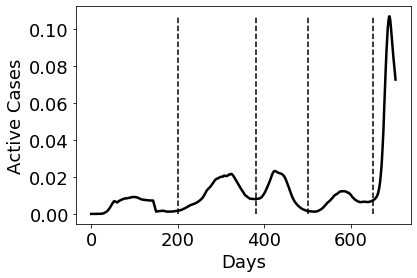}
    \caption{Proportion of daily active COVID-19 cases in Canada from the beginning of the pandemic to the end of the fifth wave.}
    \label{fig:fullCovid}
\end{figure}

In this section, we study the one-week-ahead forecast of the number of active COVID-19 cases in Canada (see Figure \ref{fig:fullCovid}) upto the fifth wave of COVID-19. Each wave is approximately given by the black vertical lines in Figure \ref{fig:fullCovid}. We notice that the first four waves are similar in terms of wavelength as well as amplitude. On the other hand, the fifth wave however is different since it is driven by a much more infectious Omicron variant and hence has a short wavelength with a higher amplitude. Therefore, we examine the forecast within the second (from day 200 to day 380)  and the fifth waves (from day 650 to day 704) of COVID-19 in Canada.

\begin{figure}[ht!]
\centering

     \includegraphics[width = 0.3\textwidth]{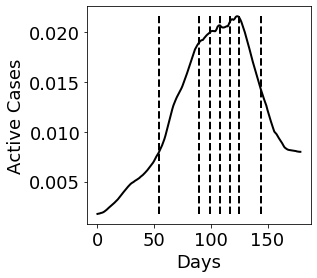}
    \includegraphics[width = 0.3\textwidth]{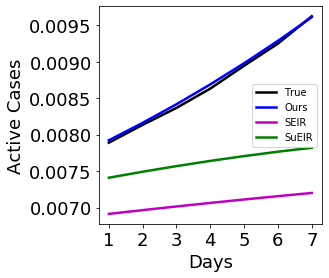}
    \includegraphics[width = 0.3\textwidth]{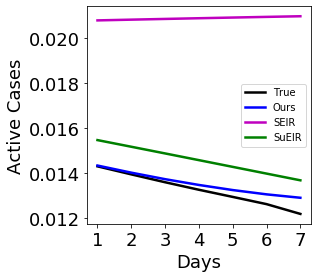}
  
    \includegraphics[width = 0.3\textwidth]{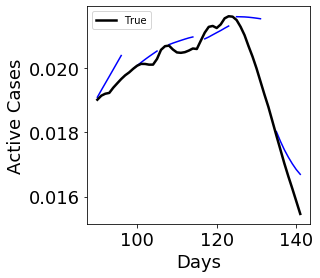}
    \includegraphics[width = 0.3\textwidth]{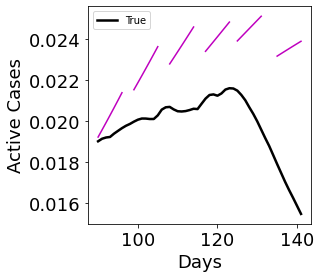}
    \includegraphics[width = 0.3\textwidth]{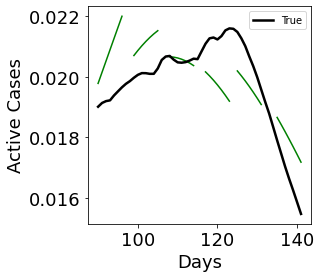}

    \caption{Results on second wave of COVID-19 in Canada. First row: Second wave from 0 upto $54^{th}$ and $144^{th}$ days out of 180 days (top left). The next two figures are the predicted values of the infectious variable $I(t)$ in the next seven days using SPADE4 (blue), SEIR model (magenta), and S$\mu$EIR model (green) correspond to those two training datasets versus ground truth (black). Second row: Prediction of $I(t)$ for next seven days around the peak of the wave using SPADE4 (blue), SEIR model (magenta), and S$\mu$EIR model (green) versus ground truth (black).  }
    \label{fig:covid2}
\end{figure}

\begin{figure}[ht!]
\centering
     \includegraphics[width = 0.3\textwidth]{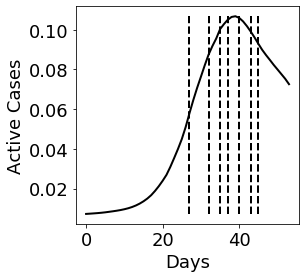}
    \includegraphics[width = 0.3\textwidth]{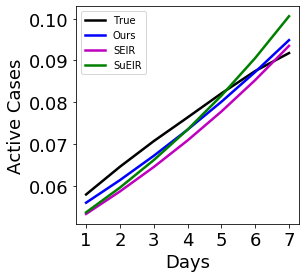}
    \includegraphics[width = 0.3\textwidth]{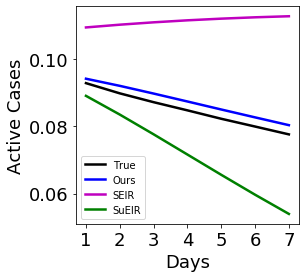}

    \includegraphics[width = 0.3\textwidth]{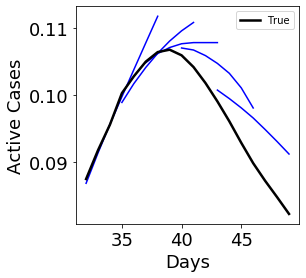}
    \includegraphics[width = 0.3\textwidth]{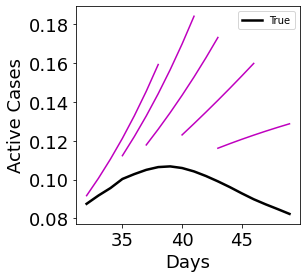}
    \includegraphics[width = 0.3\textwidth]{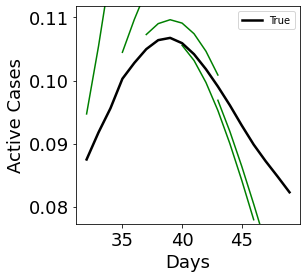}

     \caption{Results on fifth wave of COVID-19 in Canada. First row: Second wave from 0 upto $27^{th}$ and $46^{th}$ days out of 54 days (top left). The next two figures are the predicted values of the infectious variable $I(t)$ in the next seven days using SPADE4 (blue), SEIR model (magenta), and S$\mu$EIR model (green) correspond to those two training datasets versus ground truth (black). Second row: Prediction of $I(t)$ for next seven days around the peak of the wave using SPADE4 (blue), SEIR model (magenta), and S$\mu$EIR model (green) versus ground truth (black).  }
    \label{fig:covid5}
\end{figure}
The anomalies in how data were reported, such as the under-reporting of cases on weekends and the backlog cases reported later, can lead to a noisy dataset. Hence, we consider seven day average of the given data as the ground truth. For the initial conditions required by the benchmark models, we choose the initial values of the variables $I(t_0) = I_0$, $R(t_0) = 0$, where $I_0$ is the first data point in the training set. Since the initial exposed population is not known, we let the model choose the best estimate of $E(t_0)$ from $\{kI_0: k\in\{0,1,5,10,15,20,25,50,80\}\}$, which is the value of $E(t_0)$ that gives the smallest $l_2$-squared error over the training set. Finally, the initial susceptible variable is given by $S(t_0)=P-E(t_0)-I(t_0)$, where $P$ is the approximate population of Canada, $P=3.8\times10^7$. To avoid negligible values when the population $P$ is extremely large compared to the number of active cases, we normalize the training data by dividing the entire dataset by $c*P$, where $c\in (0,1]$. Here, we choose $c=0.1$. Finally, for both the second and the fifth waves, we consider the number of training data $m$ as the observed trends change, including the trends before the peak, around the peak, and after the peak. More precisely, $m$ is chosen from the set $\{54,90,99,108,117,126,135,144\}$ for the second wave and from the set $m\in\{27,32,35,38,41,43,46\}$ for the fifth wave (see the top left figures in Figure \ref{fig:covid2} and Figure \ref{fig:covid5}). Note that $m$ corresponds to the number of days starting from the starting date of a wave.

 The forecast result illustrates the advantage of our method compared to the benchmark models. More precisely, the forecasts suggested by SPADE4 (blue curves in Figures \ref{fig:covid2} and \ref{fig:covid5}) are closest to the true curve (in black) and it successfully picks up the changing nature of the curve around the peak quickly and adapts to the downward trend of the true curve. With the SEIR model, as the training data varies across before, around to after the peak (second row and top right plots in Figures \ref{fig:covid2} and \ref{fig:covid5}), it tends to move from underprediction to overprediction to an extent where it is completely away from the true curve after the peak. The S$\mu$EIR model starts with underprediction and moves on to a mix of overprediction and underprediction inconsistently without picking up the changing slope of the trajectory as the size of training data varies (second row and top right plots in Figures \ref{fig:covid2} and \ref{fig:covid5}).

\subsection{Data from Cumulative Cases of Ebola in Guinea, Zika in Giradot and influenza A/H7N9 in China}\label{sec:ebola}

In this section, we use data based on cumulative cases of Ebola in Guinea, Zika in Giradot and influenza A/H7N9 in China. For each dataset, the parameters based on the size of the dataset, population $P$ used for normalization of the dataset, preprocessing constant, size of the training set and forecast horizon have been summarized in Table \ref{tab:pars}. The initial conditions are chosen as $S(t_0) = P-E(t_0)-I(t_0)$, $I(t_0) = I_0$, $R(t_0) = 0$, where $I_0$ is the first data point in each dataset and we estimate $E(t_0)$ from $\{kI_0: k\in\{0,1,5,10,15,20,25,50,80\}\}$, which is the value of $E(t_0)$ that gives the smallest $l_2$-squared error over the training set. As in Section \ref{sec:covid}, we let the model choose the best estimate of $E(t_0)$. A convolution filter with the parameter $s=10$ (see Equation \eqref{eq:smooth}) is used on the derivative vector. We test SPADE4 on different sizes of training sets with cardinality $m$ (see Table \ref{tab:pars} for values of $m$ considered). Comparison with the benchmark method is done using one-week-forecast horizon of the proportion of cumulative cases.

\begin{figure}[ht!]
\centering

 \centering
    \includegraphics[width = 0.3\textwidth]{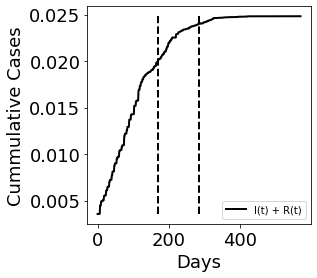}
    \includegraphics[width = 0.3\textwidth]{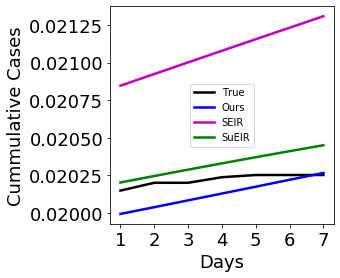}
    \includegraphics[width = 0.3\textwidth]{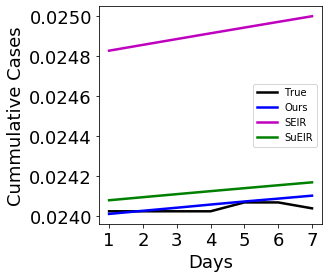}

    \includegraphics[width = 0.3\textwidth]{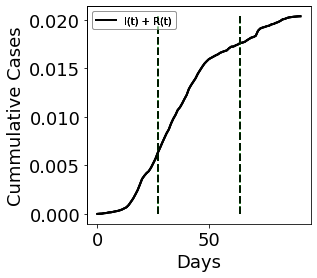}
    \includegraphics[width = 0.3\textwidth]{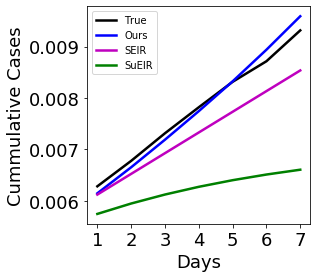}
    \includegraphics[width = 0.3\textwidth]{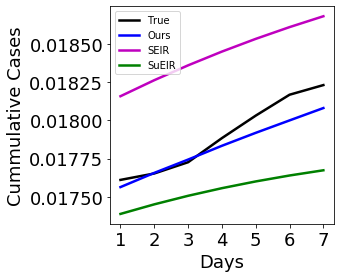}\\
    \includegraphics[width = 0.3\textwidth]{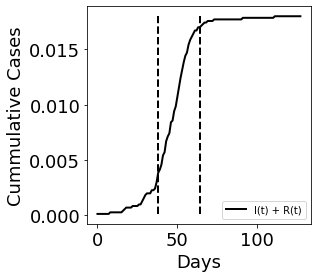}
    \includegraphics[width = 0.3\textwidth]{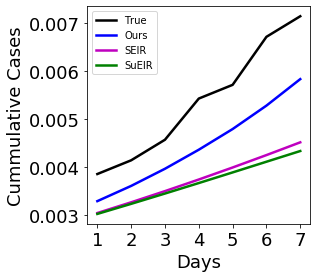}
    \includegraphics[width = 0.3\textwidth]{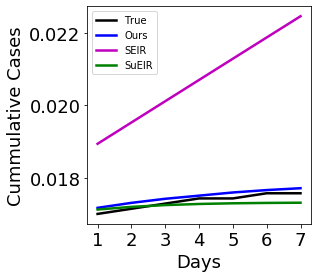}
   
    \caption{First column presents Ebola (top), Zika (center) and influenza A/H7N9 (bottom) datasets where the training data is from day 0th to the dashed vertical lines. The next two columns present the corresponding one-week ahead forecasts of our SPADE4 (in blue), SEIR (in magenta), and S$\mu$EIR (in green) versus ground truth (in black).}
    \label{fig:cummulative}
\end{figure}

\begin{table}[ht!]
\begin{center}
\begin{tabular}{ |c|c|c|c|c|c| } 
\hline
Data & Total Data  &Population& $c$ &$m$ (days) & Prediction Interval\\
\hline
\multirow{2}{10em}{Ebola Data from Guinea} & 572 & $135\times 10^6$ &$10^{-3}$&172 &173-179\\ 
&  & &&286 & 287-293 \\ 
\hline
\multirow{2}{10em}{Zika Data from Giradot} & 93 & $95 \times 10^3$ &1&27 &28-34\\ 
& & &&65 &66-72 \\
\hline
\multirow{2}{10em}{Influenza Data from China} & 128 &$7\times 10^8$ &$10^{-5}$&44 &45-51\\ 
& & &&64 &65-71\\ 
\hline

\end{tabular}
\caption{\label{tab:pars}{Parameters corresponding to datasets for Ebola, Zika and influenza A/H7N9.}}
\end{center}
\end{table}

The one-week-ahead forecasts of the Ebola, Zika and flu datasets are given in Figure \ref{fig:cummulative}. From the second column in Figure \ref{fig:cummulative}, we can see that the benchmark methods with SEIR and S$\mu$EIR have a tendency to underpredict for Zika and flu datasets when the epidemic curve has a sharp slope and the number of training points is less. For the Ebola dataset, S$\mu$EIR model learns well with less training data, however, the SEIR model overpredicts far away from the true curve. At this point, however, forecasts suggested by SPADE4 are closest to the true values. Towards the end of the curve given in the third column of Figure \ref{fig:cummulative}, an increase in training data improves the S$\mu$EIR benchmark method (green curves) predictions which are much closer to the true values, especially for Ebola and flu datasets. The SEIR method predictions are still far away irrespective of the increase in training points as the method struggles to learn the changing nature of the trajectory. Predictions of SPADE4 remain consistently close to the true values with slight improvement with an increase in the size of training data. This can be seen more clearly in Table \ref{tab:ebola} in Section \ref{sec:bt} where relative errors on the one-week-ahead horizon have been summarized. The results demonstrate that the performance of the benchmark method is dependent on the knowledge of the target model, the nature of the epidemic and the number of data points. Fitting to an incorrect target model or scarce training data can affect prediction accuracy. However, SPADE4 can predict close to the true values without any prior knowledge of the underlying model.

\subsection{Comparison with SEIR Models with Time-Varying Transmission Rate}\label{sec:bt}

\begin{figure}[ht!]
    \centering
    \includegraphics[width = 0.45\textwidth]{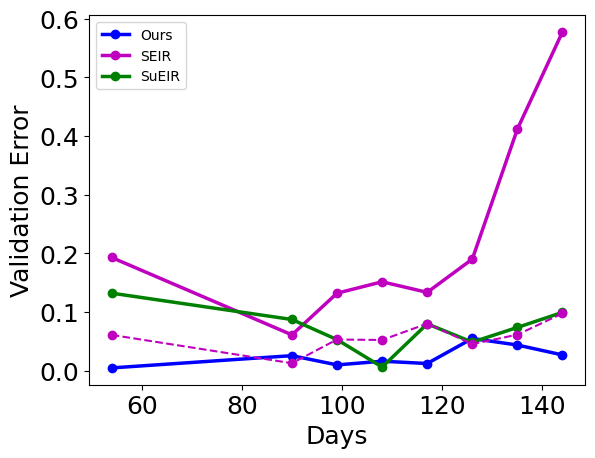}
    \includegraphics[width = 0.45\textwidth]{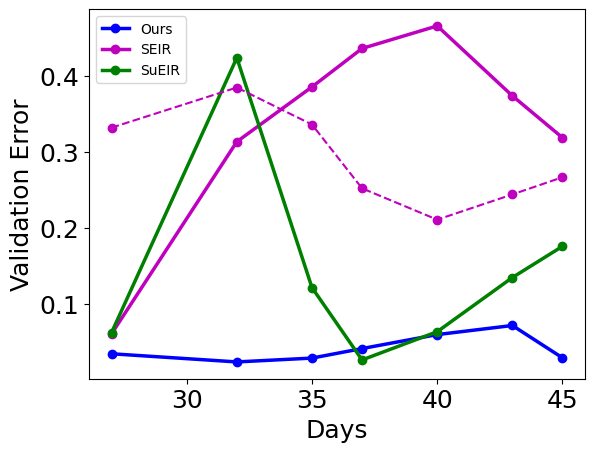}
    \caption{Validation error plot with a various number of training points for our SPADE4 (blue), SEIR (solid magenta), SEIR$_{\beta(t)}$ (broken magenta), and S$\mu$EIR model (green) models on second (left) and fifth (right) wave of COVID-19 dataset.}
    \label{valerrorcovid}
\end{figure}

In this section, we provide a comparison analysis between the performance of our SPADE4 against the time-varying transmission rate SEIR models \cite{smirnova2019forecasting,girardi2023seir}, denoted by SEIR$_{\beta(t)}$. 
In \cite{smirnova2019forecasting,girardi2023seir}, the authors present $\beta(t)$ using a fixed basis and learn $\beta(t)$ along with the other parameters of the model. In our paper, we compare with SEIR$_{\beta(t)}$,  where $\beta(t)$ has a Legendre polynomial basis representation of order $q$, i.e., given $t\in \left[a,b\right]$,
\begin{equation*}\label{eq:timevarybeta}
    \beta(t) = \sum\limits_{k=0}^q \xi_k P_k (x),
\end{equation*}
where $x = \dfrac{2t-b-a}{b-a}$ and $P_k(x)$ are Legendre polynomials of order $0\leq k\leq q$ given by
\begin{align*}
    &P_0(x) = 1,\quad P_1(x)=x, \\
    &(2k+1)P_{k+1}(x) = (k+1)xP_{k}(x) - kP_{k-1}(x).
\end{align*}
We use non-linear least squares to learn the paramaters $\gamma,\sigma$ and the weights $\xi_k$'s representing $\beta(t)$ for each $q\in\{1,2,3,4,5,6\}$ and let the model choose the best $q$ using Bayesian Information Criterion (BIC). To compare the performance of our SPADE4 with SEIR$_{\beta(t)}$, we plot the seven-day validation errors with various sizes of the training data for the second and fifth wave of COVID-19 in Canada in Figure \ref{valerrorcovid}. Additionally, we also report in Table \ref{tab:ebola} the seven-day validation errors for the datasets with cumulative data explored in Section \ref{sec:ebola}. 

From Figure \ref{valerrorcovid}, one can conclude that SPADE4 consistently outperforms SEIR, S$\mu$EIR, and SEIR$_\beta(t)$ models. Especially, the performance gap is significant for the fifth wave when the peak is sharp and the amount of available data is limited.

\begin{table}[ht!]
\begin{center}
\begin{tabular}{ |c|c|c|c|c|c|c| }
\hline
Data & $m$ & SPADE4 & SEIR&SEIR$_{\beta(t)}$ & S$\mu$EIR \\
\hline
\multirow{2}{12em}{Ebola Data from Guinea} & 172 & \textbf{0.0053} &0.0428& 0.0811 &0.0058 \\ 
&286 & \textbf{0.0012} & 0.0365&0.00497 & 0.0036 \\ 
\hline
\multirow{2}{12em}{Zika Data from Giradot} & 27 & \textbf{0.0204} &0.0638&0.1168 &0.2185\\ 
&65 & \textbf{0.0055} &0.0302& 0.0635 &0.0213  \\
\hline
\multirow{2}{12em}{Influenza Data from China} & 38 & 0.1783 &0.3178& \textbf{0.0511} &0.3359\\ 
& 64 &\textbf{0.0079} &0.2017& 0.0291 &0.0097 \\ 
\hline
\end{tabular}
\caption{\label{tab:ebola}{Relative error over one-week-ahead forecast horizon for different sizes of training data for the datasets corresponding to Ebola, Zika and influenza A/H7N9.}} 
\end{center}
\end{table}

A similar comparison has been investigated to datasets with cumulative data, including Ebola in Guinea, Zika
in Giradot, and influenza A/H7N9 in China. The seven-day forecasting errors (in terms of relative validation error) is given in Table \ref{tab:ebola}. We notice that SPADE4 performs the best in most cases except one with the influenza A/H7N9 dataset from China.

\subsection{Prediction Interval}
\begin{figure}[ht!]
    \centering
    \includegraphics[width = 0.3\textwidth]{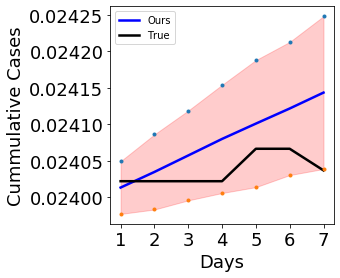}
    \includegraphics[width = 0.3\textwidth]{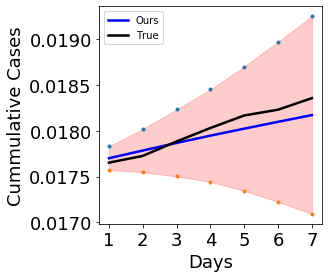}
    \includegraphics[width = 0.3\textwidth]{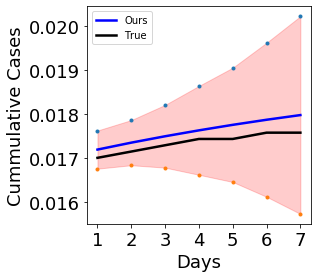}
    \caption{Prediction Interval for the Ebola dataset with $m=286$ ($m_1 = 266$ and $m_2 = 286$), Zika dataset with $m=65$ ($m_1 = 45$ and $m_2 = 65$) and influenza A/H7N9 dataset with $m=64$ ($m_1 = 44$ and $m_2 = 64$).}
    \label{fig:my_label}
\end{figure}

We propose a simple method to construct the prediction interval for SPADE4. 
The main idea is using part of the training data to estimate the  variance $\hat{\sigma}(t)$ of $\hat{I}_{ours}(t)$.
Then, the $95\%$ prediction interval is $\hat{I}_{ours}(t) \pm 1.96 \hat{\sigma}(t)$.
The detail of our construction method is outlined in Algorithm \ref{alg:coi}. 
To illustrate the performance of this method, we construct the prediction interval for the Ebola, Zika and influenza A/H7N9 datasets using the same setting as in Section \ref{sec:ebola}.
We can see that the true curve lies inside the shaded region of the 95\% prediction interval (Figure \ref{fig:my_label}).

\begin{algorithm}

\caption{Prediction Interval of SPADE4}\label{alg:coi}

\textbf{Given:} Data $\{I(t_k)\}_{k=1}^{m_2}$; Forecast window $T$; a number $m_1 < m_2 - T$ \\\

    \textbf{To find:} $95\%$ prediction interval for $\{I(t_k)\}_{k={m_2+1}}^{m_2+T}$. 
\begin{algorithmic}[1]

\Ensure

\medskip

\State{Initialize} $V = [\mathbf{v}_1,\ldots,\mathbf{v}_{m_2-m_1-T}]\in\mathbb{R}^{T\times (m_2-m_1-T)}$ and $\hat{\Sigma} = [\hat{\sigma}_1,\ldots,\hat{\sigma}_T]\in\mathbb{R}^T$.

\medskip

\For{$i=1,\ldots, m_2-m_1-T$}
\[
\begin{aligned}
&\{\hat{I}_{ours}(t_k)\}_{k=m_1+i}^{m_1+i+T} \gets \text{SPADE4}(\{I(t_k)\}_{k=1}^{m_1+i-1})\\
&\mathbf{v}_i \gets [\hat{I}_{ours}(t_{m_1+i})-I(t_{m_1+i}),\ldots,\hat{I}_{ours}(t_{m_1+i+T})-I(t_{m_1+i+T})]^T
\end{aligned}
\]
\EndFor

\medskip

\For{$j=1,\ldots ,T$}
\[\hat{\sigma}_j = \sqrt{\dfrac{1}{m_2-m_1-T}\sum\limits_{k=1}^{m_2-m_1-T} V[k,j]^2}\]
\EndFor

\medskip

\State Find $\{\hat{I}_{ours}(t_k)\}_{k=m_2+1}^{m_2+T} \gets \text{SPADE4}(\{I(t_k)\}_{k=1}^{m_2})$

\medskip

\State Prediction intervals: $[\hat{I}(t_{m_2+1})\pm \hat{\sigma}_1*1.96,\ldots,\hat{I}(t_{m_2+T})\pm \hat{\sigma}_T*1.96]$
\end{algorithmic}
\end{algorithm}

\subsection{Stability}
\begin{figure}[ht!]
    \centering
    \includegraphics[width = 0.3\textwidth]{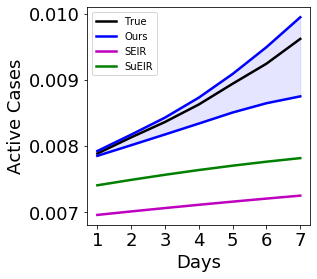}
    \includegraphics[width = 0.3\textwidth]{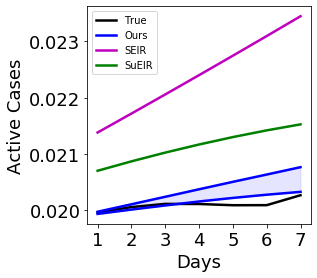}
    \includegraphics[width = 0.3\textwidth]{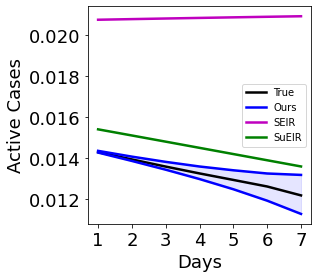}
    \caption{Predicted seven-day forecast curves using 100 runs of SPADE4 (shaded in blue) and the benchmark methods (SEIR model in magenta and S$\mu$EIR model in green) for the second wave of COVID-19 in Canada before the peak, at the peak and after the peak. }
    \label{fig:stability}
\end{figure}

Since our proposed method SPADE4 uses random features as a basis, in all our experiments above, we make use of a fixed seed to ensure the results are consistent over multiple execution of the simulations. However, in this section we provide numerical evidence of the stability of SPADE4 with respect to the random basis generated for each simulation. Figure \ref{fig:stability} gives all the possible predicted curves for the second wave of COVID-19 in Canada with different sizes of training data using SPADE4 (shaded in blue) over one hundred randomly generated basis along with the predictions of the benchmark methods. We can see that all the predictions made by SPADE4 are closest to the true curve (in black) in comparison to the other methods depicting that SPADE4 performs well consistently irrespective of the random basis generated.

\subsection{Varying Embedding Dimension}
\begin{figure}
    \centering
    \includegraphics[scale = 0.5]{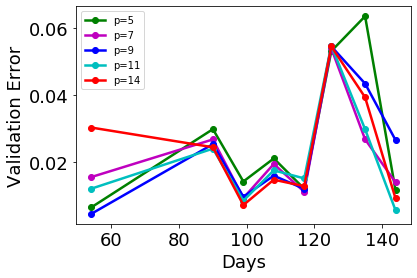}
    \caption{Validation error plot with various number of training points for SPADE4 with $p\in\{5,7,9,11,14\}$ for the second wave of COVID-19 in Canada.}
    \label{fig:emb_dim}
\end{figure}

In this section, we demonstrate the stability of our method with respect to the choice of the embedding dimension $p$. We plot the validation error for one-week ahead forecasts for the second wave of COVID-19 in Canada with varying $p$ and plot them in Figure \ref{fig:emb_dim}. We use $p=9$ in all our experiments since we follow the population division suggested by one of the most popular SEIR model in epidemiology. Since $d=4$ in SEIR and S$\mu$EIR models, we choose $p=2d+1=9$. The results in Figure \ref{fig:emb_dim} demonstrate the fact that using $p=9$ gives consistently good results across different sizes of training data. While values of $p<9$ do not perform the best, the error does improve for some cases when $p>9$. However for small input data, a large value of $p$ is not feasible since it would further reduce the training data after the use of time delay embedding map. Thus, while the embedding dimension $p$ is a hyperpaprameter worthy of further exploration, our experiments suggest that choosing $p=2d+1$ outperforms other choices when both, the size of input data and the validation error are taken into account. 

\section{Discussion and Conclusion}
Short-term forecasts of an epidemic based trajectory serve as a communication channel between the scientific community and decision makers. It helps in taking informed short-term decisions such as allocation of medical supplies, healthcare staffing needs, lockdowns and closures. We propose a delay embedding based model SPADE4 which forecasts the trajectory (daily active cases or cumulative cases) in a one-week-forecast horizon. We compare SPADE4 with benchmark methods with SEIR, SEIR$_{\beta(t)}$, and S$\mu$EIR fitting. Our experiments on simulated and real datasets show that SPADE4 can provide a reliable and comparatively accurate forecast that exceeds the performance the benchmark method. While some knowledge of the target model and initial values are required by the benchmark model, SPADE4 has no such prerequisites for its performance. Using existing theoretical results on delay embedding and sparse random feature models, we show that the choice of hyperparameters for SPADE4 is reasonable for its performance. Additionally, we also provide results to show that the predictions given by SPADE4 are stable with respect to the basis chosen as well as our choice of the embedding dimension $p$. 

Since the performance of SPADE4 is partially dependent on an accurate estimation of the derivative, the method might not give accurate results if the data is extremely noisy. Although we do use simple convolution based smoothing filters to address noisy datasets, more advanced denoising techniques might be required for higher levels of noise. Another drawback of the method is the hyperparameter choice of the embedding dimension. While the Takens' theorem gives an estimate of the embedding dimension (atleast $2d+1$), we need the number of variables present in the original system. Another direction for future work is applying SPADE4 for seasonal epidemics.

As a part of our future work, we would like to address scenarios where SPADE4 might not work very well as mentioned above.  Further direction of this work involves addressing the scenarios where SPADE4 may not perform very well as highlighted above.

\section*{Data Availability Statements}
The code is available at 

\url{https://github.com/esha-saha/spade4} and data sources are
\begin{itemize}
\item COVID-19 in Canada: 

\url{https://health-infobase.canada.ca/covid-19/epidemiological-summary-covid-19-cases.html}
\item Ebola in Guinea: 

\url{https://www.kaggle.com/datasets/imdevskp/ebola-outbreak-20142016-complete-dataset}
\item Zika virus in Giradot: 

\cite{rojas2016epidemiology}
\item Influenza A/H7N9 in China: 

\url{https://datadryad.org/stash/dataset/doi:10.5061/dryad.2g43n}
\end{itemize}

\section*{Acknowledgement}

LSTH was supported by the Canada Research Chairs program, the NSERC Discovery Grant RGPIN-2018-05447, and the NSERC Discovery Launch Supplement DGECR-2018-00181. E. Saha and G. Tran were supported by the NSERC Discovery Grant and the NSERC Discovery Launch Supplement.

\pagebreak
\section*{Appendix A}
\subsection*{Results on Simulated Data with $2\%$ Noise}\label{secA1}
\begin{figure}[h!]
\centering
     \includegraphics[width = 0.3\textwidth]{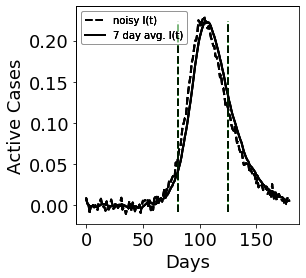}
    \includegraphics[width = 0.3\textwidth]{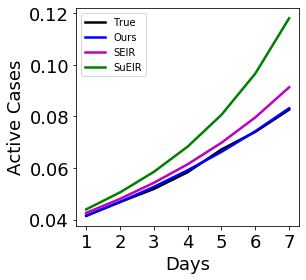}
    \includegraphics[width = 0.3\textwidth]{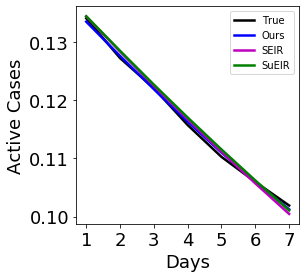}\\
    \includegraphics[width = 0.3\textwidth]{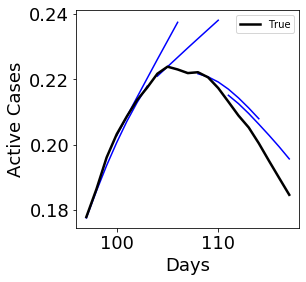}
    \includegraphics[width = 0.3\textwidth]{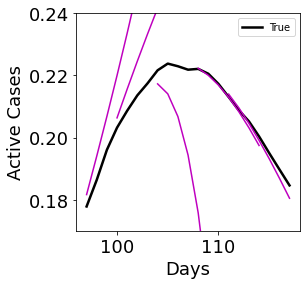}
    \includegraphics[width = 0.3\textwidth]{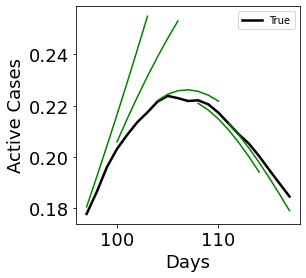}

     \caption{Results on simulated data with $2\%$ noise added to input data. First row: Noisy training datasets from 0 upto $81^{st}$ and $125^{th}$ days out of 180 days (top left). The next two figures are the predicted values of the infectious variable $I(t)$ in the next seven days using SPADE4 (blue), SEIR model (magenta), and S$\mu$EIR model (green) correspond to those two training datasets versus ground truth (black). Second row: Prediction of $I(t)$ for next seven days around the peak of the wave using SPADE4 (blue), SEIR model (magenta), and S$\mu$EIR model (green) versus ground truth (black).}
    \label{fig:SimulNoisy2}
\end{figure}

\bibliographystyle{abbrv}
\bibliography{references}{}
\end{document}